\documentclass[11pt,a4paper]{article}
\usepackage[hyperref]{acl2021}

\usepackage{times}
\usepackage{latexsym}

\usepackage[T1]{fontenc}
\usepackage[utf8]{inputenc}

\usepackage{microtype}
\usepackage{graphicx}
\usepackage{latexsym}
\usepackage{booktabs}
\usepackage{amsmath}
\usepackage{multirow}
\usepackage{amssymb}
\usepackage{booktabs}
\usepackage{url}
\aclfinalcopy
\title{
Evidence-based Factual Error Correction
}
\renewcommand\footnotemark{}
\author{James Thorne \\ %\thanks{Preprint: work in progress} \\
Department of Computer Science\\
  University of Cambridge \\
  {\tt jt719@cam.ac.uk} \\\And
  Andreas Vlachos \\
  Department of Computer Science \\
  University of Cambridge \\
  {\tt av308@cam.ac.uk} \\}

\date{}
\begin{document}
\maketitle

\renewcommand\footnotemark{\arabic}
\begin{abstract}
This paper introduces the task of factual error correction: performing edits to a claim so that the generated rewrite is better supported by evidence. 
This extends the well-studied task of fact verification by providing a mechanism to correct written texts that are refuted or only partially supported by evidence.
We demonstrate that it is feasible to train factual error correction systems from existing fact checking datasets which only contain labeled claims accompanied by evidence, but not the correction. 
We achieve this by employing a two-stage distant supervision approach that incorporates evidence into masked claims when generating corrections.
Our approach, based on the T5 transformer and using retrieved evidence, achieved better results than existing work which used a pointer copy network and gold evidence, producing accurate factual error corrections for 5x more instances in human evaluation and a .125 increase in SARI score.
The evaluation is conducted on a dataset of 65,000 instances based on a recent fact verification shared task and we release it to enable further work on the task.\footnote{\url{https://github.com/j6mes/2021-acl-factual-error-correction}}
\end{abstract}

\section{Introduction}
Fact verification is the task of predicting whether claims are true or false using evidence. With the availability of a number of resources \citep{Wang2017a,Karadzhov2017FullySources,Thorne2018a,Augenstein2019,Wadden2020FactClaims}, the task has attracted significant attention and spawned the development of new models, architectures and approaches. 
With potentially sensitive applications, recent works have focused on building explainable  variants of fact checking \citep{Atanasova2020GeneratingExplanations, Stammbach2020,Kotonya2020}. Exposing the evidence source and decision making process may help the reader uncover subtle issues that cause automated systems to fail. 
Additionally, using such evidence to continuously update news articles as facts change forms part of the vision outlined by \citet{Cohen2011} for automated newsrooms.
%Understanding how to make changes to documents based on updated evidence would help fulfill this vision.

%For the FEVER shared task \citep{Thorne2018a}, a large number of systems are a pipeline of information retrieval and classification: relevant evidence is selected from a corpus of texts and used to classify whether the claim is \textsc{Supported} or \textsc{Refuted} by it.

\begin{figure}
    \centering
    \includegraphics[width=0.9\linewidth]{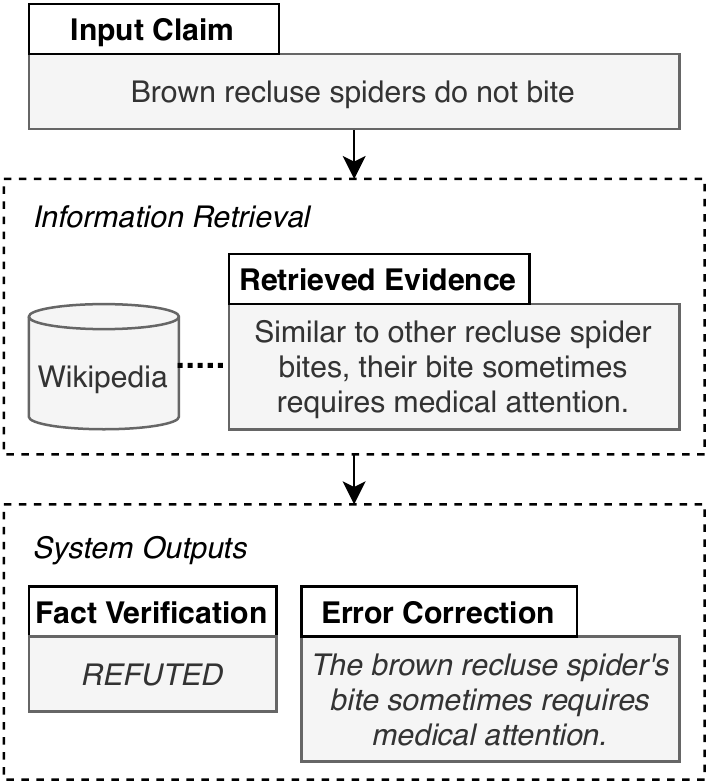}
    \caption{Factual Error Correction uses evidence to make corrections to claims, in contrast to fact verification, which instead classifies the veracity of the claim.}

    \label{fig:intro_example}
\end{figure}

In this paper, we propose \emph{Factual Error Correction}, as an explainable alternative for fact verification. 
Rather than merely assigning a truth label, possibly accompanied by evidence, our goal is to rewrite claims so that they are better supported by the retrieved evidence. 
For example, in Figure~\ref{fig:intro_example}, a claim that would be \textsc{Refuted} by the evidence using a fact verification system is rewritten so that it becomes supported by evidence retrieved from Wikipedia.
%We extend previous work by requiring that the evidence is retrieved from a collection of documents representing ground truth facts (in our case Wikipedia).}% The approach we take, that does not require additional training data, is to mask out conflicting tokens prior to making the correction.
This work extends fact guided sentence modification \citep{Shah2019}, which uses short factoid claims to introduce changes to Wikipedia passages.
%, which uses masking and self-supervision to obviate the need for additional training data, 
%Whereas \citet{Shah2019} %the previous work 
However, they
assume that the claim and Wikipedia text are always incongruous and require a meaning-altering change, our proposal makes no assumptions over the veracity,
and is applicable to
claims both supported and refuted by evidence.
%We apply our pipeline to  claims that supported by evidence, as well as those which are refuted. 
%This summarizes the evidence with respect to the claim, helping to resolve half truths, and better informs the end-user of the decision making process, regardless of label. 
Additionally, we incorporate a retrieval component to select evidence for a given claim from a corpus (in our case, Wikipedia) rather than requiring gold standard evidence to be explicitly provided.

%While this appears similar to Grammatical Error Correction \citep[GEC]{Knight1994AutomatedDocuments, Ng2014}, with the common goal of providing feedback to an end-user, However, unlike GEC, all our modifications are performed by incorporating external evidence.

%An outline of our proposal is illustrated in Figure~\ref{fig:my_label} 

%This allows highlighting where there is disagreement between the claim and evidence. However, unlike GEC, all our modifications are performed by incorporating external evidence.
%Furthermore, in \emph{factual} error correction, the generated corrections both need to remove the error and be supported by the evidence. 
%Generating corrections provides a mechanism that helps explain the decision making process better than simply assigning a label or highlighting tokens. While generating post-hoc \emph{black-box} model explanations \citep{Ribeiro2016, Ribeiro2018} or interpreting attention \citep{Jain2019,Thorne2019} are informative, they better describe the behaviour of a model -- rather than actionable measures the user can take.
%Corrections inform the end user where the misinformation was introduced, loosely following work on building inherently explainable models (such as \citet{Park2018MultimodalEvidence, Camburu2018}).

A challenge for factual error correction is the lack of datasets consisting of claims paired with their corrections. 
However, with recent developments in fact checking, there is an abundance of new datasets consisting of claims paired with evidence.
To address this data scarcity, we make use of distant supervision to incorporate retrieved evidence into generating the corrections.
%To characterise the challenges in factual error correction, we develop a distantly supervised baseline, trained using FEVER \citep{Thorne2018a}, combining retrieved evidence from DPR \citep{Karpukhin2020} and a T5 seq2seq transformer \citep{Raffel2020}. This system attains a SARI score of 0.XXXX. 
%We compare this to the system from \citet{Shah2019}, which attained a SARI score of 0.XXXX.

We release a dataset of 65,000 claims, containing the intermediate annotations from FEVER \citep{Thorne2018a}. These consist of factoid sentences that were used to construct the supported and refuted claims in the dataset, and use these as reference targets for automated evaluation.% These annotations describe how facts from Wikipedia were mutated to construct claims. we use the \emph{unmutated} facts as reference corrections to automate evaluation of our distantly-supervised systems as well as train a fully supervised baseline indicative of ceiling performance, which attains a SARI of 0.XXXX. 
We further verify the findings through a final round of annotation using human raters. Our evaluation finds high correlation between manual scores and the SARI metric \citep{xu-etal-2016-optimizing} and our best performing distantly-supervised system generated corrected claims for 24\% of instances when using retrieved evidence, with a SARI Final score of .419. A fully-supervised system with gold evidence generated corrections for 69\% of instances, indicating plenty of opportunities for future work to extend our contributions.

\section{Related Work}
\label{sec:related}
A number of related works offer methods to make corrections to sentences. However, their use of external information differs. This can be placed on a continuum from only using the knowledge captured during language model pre-training, to conditioning generation based on a context sentence. We briefly outline key methods and approaches below.

Grammatical Error Correction (GEC) \citep{Knight1994AutomatedDocuments, han-etal-2010-using, Ng2014} is the task of making \emph{meaning-preserving} changes to sentences such that grammatical errors made by language learners are removed. No external information is required as the sentence is undergoing a surface-level transformation where the (intended) semantic content of the sentence should remain unchanged. 
%Recent works model GEC as a sequence-to-sequence task
%han-etal-2010-using
%\cite{Yuan2016}, similar to machine translation.

%is presented in It is a requirement that the generated corrections both remove the error from the original claim and be supported by the evidence in the generated correction. 

In contrast, the semantic content of sentences undergoing \emph{factual} error correction will be altered, if needed, to better align the meaning with ground truth evidence. 
\citet{Shah2019} make meaning-altering updates to sentences in Wikipedia in a two step process that does not require reference corrections in training: salient tokens are masked and a corrector conditionally replaces the masks with ground truth evidence. In this approach, token salience is predicted by querying a model that is trained to perform fact verification for a claim against evidence. 
\citet{Cao2020FactualModels} generate corrections as a post-editing step for outputs from abstractive summarization so that they are consistent with the source text.  Their approach uses a sequence-to-sequence model trained to restore artificially generated corruptions of a reference summary. 

One potential way to introduce knowledge is to use information stored in the parameters of large-scale pre-trained language models \citep{Petroni2019}. 
The language model can be used recover tokens responsible for causing factual errors that are masked out as a variant of cloze-style evaluation \citep{Taylor1953ClozeReadability}.
While such approaches have been employed for
fact verification
%(for example, 
\citep{Lee2020LanguageCheckers},
%used a BERT language %model \citep{Devlin2019}), 
these approaches share the following limitations. 
%When masked language models are used to correct claims, the underlying task is to predict the most likely token to replace a mask. 
Without explicit control \citep{nie-etal-2019-simple}, the most likely token when decoded may not be factually accurate, or supported by the retrieved evidence, commonly referred to as a hallucination \citep{rohrbach-etal-2018-object,Zhou2020DetectingGeneration}.
Furthermore, \emph{even if} the information stored within language model parameters could be reliably retrieved for factual error correction, facts change over time and the need to obtain information from up-to-date sources becomes greater as the state of the world diverges from the information captured within the model parameters.
Recent language models augmented with a retrieval component such as 
REALM \citep{Guu2020} and RAG \citep{Lewis2020} could be applied, however, task-specific fine-tuning would still be required to condition the generation based on the factual error to mitigate hallucination. %Our approach to masked correction, can be viewed as a  %However,  on a factual error is an open problem. 

\section{Task Definition}
\label{sec:task}

\paragraph{Training} 
\begin{figure*}[!t]
    \centering
    \includegraphics[width=\linewidth]{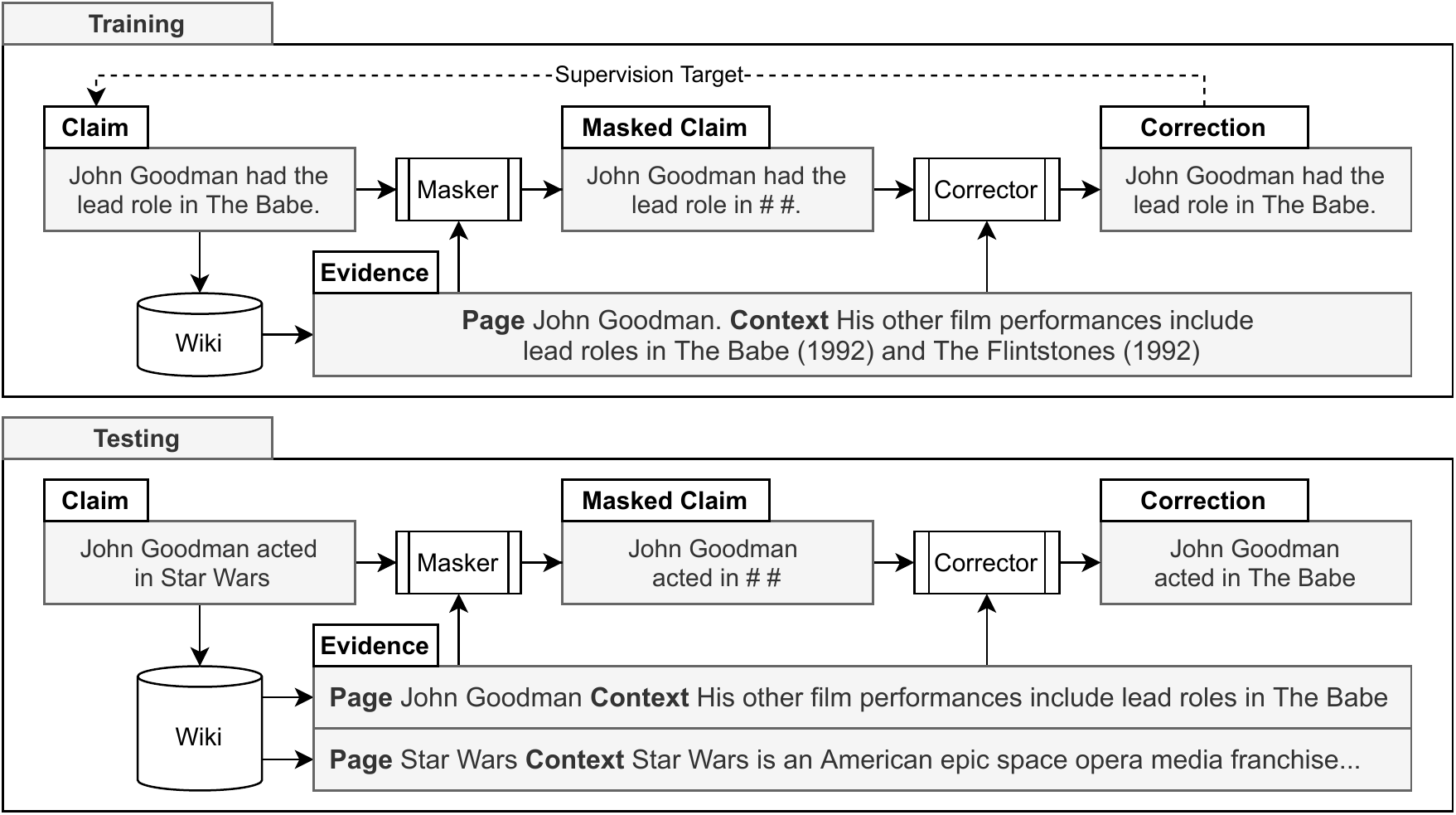}
    \caption{The corrector is trained to reconstruct masked claims, conditioned on retrieved evidence, indicated by the dashed arrow. At test time, the corrector is able to incorporate new facts from the evidence to generate corrections.} %The supervision  for supervision is the original claim input (dashed line).}
    \label{fig:training}
\end{figure*}

Let a claim $c$ be the input sentence undergoing correction to yield $c'$. The correction requires incorporating knowledge from retrieved evidence $E(c)$ such that $c'$ is supported by this evidence, {$E(c)~\vDash~c'$}. 
%The example in Figure~\ref{fig:intro_example} demonstrates how the evidence is retrieved from Wikipedia and integrated into the claim changing the label from refuted to supported.
Factual error correction is subject to the following 3 requirements:
%While in \emph{fact verification} the task is to assign a label to the claim using this retrieved evidence, \emph{error correction} differs as the evidence must be used to remove the error from the claim and rewrite it so that the correction is supported by the evidence. This is outlined by the following 3 requirements:
% al error correction is the task of making meaning-altering changes to a claim $c''$ given evidence retrieved from a corpus of documents. %The example illustrated in Figure~\ref{fig:intro_example} demonstrates how the retrieved evidence is retrieved from Wikipedia and integrated into the claim changing the label from refuted to supported.
%For a given claim, a factual error correction system must first retrieve evidence that \textsc{Supports} or \textsc{Refutes} it from a corpus of ground truth documents.

% for grammaticality, non-redundancy, clarity, focus, and coherence.
%summarized with respect to the error in the input claim. Retrieval of multiple evidence passages 
%FinallyThe corrections are assessed against criteria that consider the intelligibility in isolation, veracity using retrieved evidence, and whether any error present in the input is removed.
%Systems will be evaluated against the following requirements: 

% standalone
\paragraph{R1 - Intelligible} Similar to other language generation tasks, 
%surveyed by \citet{Celikyilmaz2020EvaluationSurvey},
our first requirement is that generated outputs are fluent and intelligible. They must be free of grammatical mistakes and the meaning must be understandable without the aid of additional context or evidence so that their factual correctness can be assessed. 

% Similar to query focused summarization \citep{Dang2005Overview2005}, retrieved evidence passages are combined with respect to the input claim to generate a short text output. While the tasks serve different purposes, there are common requirements and criteria for the generated outputs: notably with respect to grammaticality and focus (focus is also discussed in R3).  
%In our task, the corrections must be in the form of an intelligible fact.

%compared to evidence
\paragraph{R2 - Supported by Evidence} The generated correction must be supported by the retrieved evidence. This property follows from previous work \citep{Thorne2018a} % on fact verification %which mandates that person would assume that the claim is true given the information provided in the evidence. 
and also requires models to condition generation on the retrieved evidence -- penalizing models that hallucinate \citep{Holtzman2019TheDegeneration}.

%This constrains the output in two ways: the system should not be hallucinating new information, and any parts of the claim refuted by the evidence should be corrected. 
%We take the definition of Supported from the FEVER task \citep{Thorne2018a} where a reasonable person would assume that the claim is true given the information provided in the evidence.

%compared to original
\paragraph{R3 - Error correction} Specific to our task, the corrections should be targeted to the errors present in the inputted claim.  While this, in part, can be assessed by R2 we need to compare the correction to the inputted claim to ensure the output is not introducing new unrelated information. For example, an erroneous claim: \emph{France is in South America} could be supported by evidence if it were rewritten as \emph{France is a republic}. However, the desired correction should instead state \emph{France is in Europe}.

\section{Task Decomposition}

The choice of supervision for the error correction system influences the task decomposition. For example, with full supervision, the system can be constructed with an information retrieval module and a sequence-to-sequence module that conditionally generates a correction given the claim and evidence.
However, large datasets of claims paired with corrections are not available.
The absence of full supervision requires that we  distantly-supervise our systems using \emph{fact verification} datasets, which are an abundant resource.
Fact verification datasets contain claims labeled with evidence but do not contain corrections. 
%There are (which would ideally consist of claims paired with their corrections) required for supervising training.
%:We existing \emph{fact verification} datasets provide a useful resource that we exploit. 
With this resource, we propose a task decomposition that generated corrections by training models to reconstruct claims with masked tokens using retrieved evidence.

% without the need for supervision over the target correction in the training data. This approach is outlined in
% The .
% which consist of claims annotated with evidence needed to supported or refuted them.
% %it is possible to train a corrector to incorporate retrieved evidence using

%  We conditionally mask tokens from the claim and replace these, also conditioning their replacements using evidence.  Section~\ref{sec:task:distant}

\subsection{Distantly-supervised corrections}
\label{sec:task:distant}
% Given the retrieved evidence, a system must remove errors from the claim and generate its correction. 
% It is possible to train a seq2seq model without the need for claims paired with their correction using distant-supervision from a fact verification system.
% The goal is to incorporate new ground truth facts from the retrieved evidence into a masked version of the claim, assuming that tokens that contribute to factual inaccuracy can be masked out from the claim. 
% We train our factual error correction system to recover masked tokens by conditioning generation from supporting evidence.
%Assuming erroneous tokens can be masked out, corrections can be made by conditioning token recovery on the retrieved evidence

\paragraph{Test time} Corrections are generated by a two-stage process, illustrated in Figure~\ref{fig:training}. Tokens from the claim, $c$, are first masked, yielding $\tilde{c}$, and then input to the corrector $c'~=~Corr(\tilde{c}, E(c))$.
% ims, $\tilde{c}~=~Mask(c,E(c))$, are generated conditioned on retrieved evidence $E(c)$. 
The masker, $\tilde{c}~=~Mask(c,E(c))$, replaces a subset of tokens in the claim with a blank placeholder, conditioned on $E(c)$. Its purpose is to remove tokens that are salient to the claim being supported or refuted by the evidence. Using the masked claim, $\tilde{c}$, the corrector replaces the blank placeholders with tokens conditionally generated using retrieved evidence.
To correct errors, evidence refuting a claim ($E(c)\nvDash c$) conditions generation of a correction supported by it $E(c)\vDash c'$.  This extends the protocol \citet{Shah2019} by conditioning both the masker and corrector with multiple retrieved evidence sentences, rather than a single gold factoid.

% It is assumed that at test time, The two components, the masker and corrector The corrector is traiend to condition generation g
% The corrector $Corr$, works under the assumption that at training, the model  tokens in the claim that cause it to be contradicted by the evidence can be appropriately masked out and replaced so that the correction is consistent with the retrieved evidence.
% We first summarize the differences between training and test time for this corrector module, before discussing different masking strategies.
% Using the approach outlined above, we have a system that is trained to recover masked tokens in a claim conditioned on retrieved evidence.
% During training, the retrieved evidence would be supporting the claim. However, at test time, we apply the system to claims refuted by the evidence. 

% Given a claim, $c$, the input to the corrector is a masked version of the claim $\tilde{c}$ and retrieved evidence $E$. 
%Without labeled data, describing a reference correction, the corrector module must instead be trained with a proxy objective as illustrated in Figure~\ref{fig:training}.
% 

\paragraph{Training the corrector}
%In the absence of target corrections to supervise training, we distantly supervise the corrector using claims from a fact checking dataset. 
Similar to masked language modeling, the training objective is to generate the input claim $c'=c$ conditioned on the masked claim $\tilde{c}$ and evidence $E(c)$. %, prior to masking (illustrated by the dashed arrow in Figure~\ref{fig:training}). Explicitly, this is maximizing the likelihood of $P(c'=c|\tilde{c},E(c))$.
%While the knowledge captured by the parameters of masked language models can be used to predict suitable tokens (as is the training objective of the BERT language model \citet{Devlin2019}) our application differs as the generated tokens must be conditioned on the retrieved evidence to allow new information to be incorporated into claims.
%The hypothesis is that conditioning generation on evidence will allow a different corrections to be generated based on different evidence.
By training the model to generate the input claim, we expect the model to generate the input claim only if it was in complete agreement with the evidence
 (assuming the masking and the evidence are correct). Otherwise, the generated correction will contain evidence pertinent to the correcting the masked claim, which enables us to generate corrections satisfying requirements R2 and R3.

% With suitable training, atgenerate the correction $c'$ such that $E \vDash c'$ by conditioning the generation of the correction on the retrieved evidence. 

% are incorporated during the correction.
%While the corrector is trained to recover the original meaning, it is the the conditioning on evidence that means that facts consistent with the evidence are incorporated during the correction.

% factual error correction can be performed  we can apply this, the  generate corrections conditioned 
% The goal of the corrector is to incorporate retrieved evidence $E \subset \mathcal{E}$ into the masked so that the generated correction $c'$ is more strongly supported by this evidence. 
% For claims that are refuted by evidence, the task is to generate a corrected version of the claim $c'$ using the evidence, that both removes the error and becomes supported by the evidence.
% For claims that are weakly supported by evidence as in the case of half-truths or paraphrases, the generated correction should be more strongly supported by the evidence.
% During training, mask

% replace masked tokens through masking tokens from thewithout the need for additional data. 
% {\color{red} the corrector has 2 different modes, when training}
% We present a generalization of definition of the masker component, to better align it with existing work on generating token-level explanations.

%using a single simple sentence describing the fact to be introduced to perform the update, 
%we use a set of sentences from Wikipedia to form the evidence. 

\paragraph{Masker} When applied to factual error correction, masking the tokens from the claim acts as a proxy to which tokens need to be removed to correct an error. 
 Parallels can be drawn between masking and generating token-level explanations. We briefly summarize common approaches to generating explanations in Section~\ref{sec:model:masks}. %Our approach assumes that the masker is independent to the corrector.

\section{Model}

\subsection{Evidence retrieval}
We use GENRE \citep{DeCao2020AutoregressiveRetrieval} and Dense Passage Retrieval \citep{Karpukhin2020} together to retrieve evidence for claims $E(c)$. Both have shown success for a number of language understanding tasks over Wikipedia \cite{Petroni2020KILT:Tasks}. %This evidence conditions both the downstream masker and error correction components.
GENRE is a pre-trained seq2seq model, trained to predict a Wikipedia page name for a claim.  DPR encodes fixed length passages from Wikipedia into vectors using a BERT encoder to build a static index. At test-time, the claim is encoded and the most-similar passages are returned using an inner-product search. We return the top-$k$ passages returned by DPR from pages predicted by GENRE.
% \begin{equation}
% \label{eqn:dpr}
% \begin{aligned}
%     \text{DPR}(c) &= \textrm{top-k}_{d_i \in \mathcal{D}} (f_q(c)^\top f_p(d_i)) \\
%     E(c) &= \text{DPR}(c) \cap \bigcup_{p \in \text{GENRE}(c)} \text{Wiki}(p)
% \end{aligned}
% \end{equation}

%DPR has shown success for a number of language understanding tasks over Wikipedia~\cite{Petroni2020KILT:Tasks}, including the related task of fact verification.

%, even demonstrating a higher accuracy some purpose built systems for the FEVER fact verification task.

\subsection{Token-level explanations as masks}
\label{sec:masks}
\label{sec:model:masks}
At test time, the purpose of the masker is to selectively remove tokens that contribute to the factual errors within a claim. 
%This broadly follows recent work on generating token-level explanations for NLI \citep{Thorne2019a} where salient tokens can be masked out, allowing new evidence to be introduced from the retriever during correction. 
We study how the choice of masker influences the quality of corrections.
This considers varying levels of access to model information and different run-time complexity. Both the black- and white-box methods, outlined below, require querying a model trained to classify the veracity of claims given evidence whereas the the language model masker and baselines do not.
%Hyper-parameter choices are discussed in Section~\ref{sec:impl}.
%Each masker we use requires different access to model information and differing run-time complexity outlined below.

%There is a trade-off between the time taken to generate the masks, and how the masked tokens agree with human judgements of token importance \citep{Ribeiro2016}. 

\paragraph{Black-box masker} We evaluate perturbing the input to a classifier that is trained to predict the veracity of a claim given evidence. We use LIME \citep{Ribeiro2016}, a diagnostic that trains a locally linear
model to score the importance of input features (in our case, tokens in the claim) with respect to the predicted labels.
The model under test is a BERT classifier where evidence and the claim are concatenated in the input.
This is referred to as \emph{black-box} because the model does not undergo modification and no information about internal values or states is exposed. 

% Tokens are masked from the claim using predictions from a sentence pair classification model trained on instances from the FEVER dataset. This model is a fine-tuned BERT classifier that is representative of the state of the art for the FEVER shared task \citep{Thorne2018b}. We use LIME \cite{Ribeiro2016} to generate explanations of which tokens from the claim are salient to a sentence pair classifier trained on claims and evidence. This method perturbs instances by masking tokens and uses the change in labels to train a linear classifier that weights the importance of each token.

\paragraph{White-box masker}
In contrast, to obtain \emph{white-box} model explanations, the model has undergone modification to expose internal information.  
We use the Neutrality Masker from \citep{Shah2019} to predict which tokens, when masked, are likely to cause a label flip from supports or refuted to not enough information. %for a fact verification classifier.
This masker exposes encoded input of an ESIM classifier \citep{Chen2016}, and adds a linear classifier over the hidden states to predict per-token masking probability. At test time, masks can be generated through a single query to the model (unlike LIME in the black-box masker which requires multiple queries to the model), however this requires an additional step to train, using predictions from the classifier as signal.%. The signal for this linear classifier comes The signal for the training is coming from the pre-trained (now fixed) classifier.to self-supervise. 

\paragraph{Language model masker} We evaluate whether it is possible to generate %meaningful 
masks without the need for a fact verification model. We use a BERT pre-trained language model \citep{Devlin2019} to measure the surprisal of tokens in the claim. 
Our intuition is to identify tokens which introduce misinformation under the hypothesis that the world knowledge \cite{Petroni2019} captured in re-training would assign lower probabilities to tokens contradictory to the world state. This language model has no additional task-specific fine-tuning. We independently predict the cross-entropy for each token under a masked language modelling objective using BERT and return the top-k tokens.% We pass the masked the claim through the language model and greedily decode one tooke,  iteratively make predictions to the model with one token masked at a time and return the top-k with the highest per-token cross-entropy. 

\paragraph{Baselines} We additionally consider two simple baseline maskers: \textbf{random} masking of a subset of tokens and also a \textbf{heuristic} method of masking tokens which are not in common between the claim and the retrieved evidence. 

%this classifier is  to an ESIM model \citep{Chen2016} rather than interactively querying the model.
%Other methods for ``white-box'' model explanations expose internal information from the models such as attention. However attention alone may be insufficient to generate quality explanations \citep{Jain2019}. 

% On one extreme, tokens could be randomly masked, following a similar training strategy to BERT \citep{Devlin2019} or the unsupervised objective in T5 \citep{Raffel2020} pre-trained language models. 
% Generating large quantities of training data is possible without query a model to generate explanations. 
%To overcome this limitation, the authors fine-tune the attention mechanism with hand-crafted constraints to better align the generated masks with the saliency of tokens. 
% \begin{description}
%     \item[White-box masker] We use the masker from \citet{Shah2019}.
%     Tokens are masked using a linear classifier over the encoded inputs. This classifier is supervised to predict which tokens, when masked would cause a label flip from the \textsc{Supports} or \textsc{Refutes} labels to \textsc{NotEnoughInfo} with additional regularization to control mask size. 
%    
%end{description}

\subsection{Corrections}
We train an encoder-decoder transformer model to generate corrections from masked claims and evidence. 
Our model uses a pre-trained T5 transformer \citep{Raffel2020} which we fine-tune with the distant supervision protocol described in Section~\ref{sec:task:distant}. 
This model jointly encodes the masked claim and evidence by concatenating these two inputs in the input. %$Enc$. % Tokens are decoded left-to-right conditioned on previous outputs, $c_{1:t-1}'$, with decoder $Dec$.
%\begin{equation}
%\begin{aligned}
%Z &= Enc(Concat(\mask{c},E(c)) \\
%p(c_t | \tilde{c}, E(c), c'_{1:t-1}) &\propto %Dec(Z,c'_{1:t-1})
%\end{aligned}
%\end{equation}

% pretrained with a sequence to sequence masked language modelling objective.
% This pre-training objective is very similar to the task of correction, although there is no conditioning on evidence.
% To condition on evidence, we additionally fine-tune the T5 transformer to recover masked tokens
%We first consider methods which do not require additional labeled data based off recent works before presenting language modelling baselines and a fully supervised system highlighting ceiling performance.
%Changes are made to claims regardless of their veracity.

%\paragraph{Comparison to existing work}
We also compare against a baseline model from a related task of fact guided sentence modification  \citep{Shah2019} which uses a pointer generator network \citep{See2017GetNetworks}. Unlike our model, which captures long-range dependencies between claim and evidence through the transformer self-attention \citep{Vaswani2017}, the baseline independently encodes the evidence and masked claim using LSTMs \citep{Hochreiter1997} before decoding using a pointer-copy mechanism.
% Rather than independently encode the inputs, the transformer model we use (outlined above) jointly encodes the claim and evidence to condition generation of the generation of the correction.
%We provide direct comparison between this approach and our conditional generation in the evaluation in Section~\ref{sec:results}.
%During training, the corrector is trained to recover the masked tokens using the evidence -- illustrated in Figure~\ref{fig:training}. %At test time, different evidence is used (for example, retrieved evidence that \textsc{Refutes} a claim), the hypothesis is that conditioning on this different evidence would cause the tokens replacing the masked tokens to be supported by the new evidence -- correcting any factual errors.
%The model is trained for up to 8 epochs, the weights from the model which has the highest Rouge2 score on the validation set is retained for evaluation.
%Each masked token is replaced with a unique, and ordered masking token {\tt <extra\_id\_i>} where $i$ corresponds to the index of the masked token (out of all masked tokens).
% We evaluate the quality of corrections trained using different maskers as this is likely to impact the effectiveness of the generated correction.

%\paragraph{Language Models as Correctors?}
In %a final experiment, evaluating
order to evaluate
the impact of conditioning on evidence, we decode tokens from masked claims using a language model without fine-tuning or conditioning, similar to the Language Models as Knowledge Bases hypothesis introduced by \citet{Petroni2019}. This would consider correcting claims using the implicit knowledge stored within the model parameters rather than using external evidence.

\section{Data}
\label{sec:data}
%Distant supervision of factual error correction is possible using instances consisting of claims paired with evidence to incorporate new information from. 
We make use of FEVER \citep{Thorne2018a}, a commonly used fact verification dataset, as the basis for our experiments.
FEVER is one of the largest resources consisting of claims paired with evidence from Wikipedia. There are 185k instances with corresponding evidence sentences and a label as to whether the claim is \textsc{Supported} or \textsc{Refuted} by it.  
Claims where no information could be found are labeled as \textsc{NotEnoughInfo}.

To comprehensively evaluate the corrections generated % by error correction systems,
manual evaluation is required. 
However, this is expensive and not suitable for system %debugging and 
development and hyper-parameter optimization.
To automate system evaluation or to train a seq2seq model with full supervision, a reference ``gold standard'' correction is also required. 
For this, we release %intermediate
annotations from the FEVER shared task as follows.
The claims in FEVER were generated in a two-stage process: annotators extracted facts from Wikipedia and then performed meaning altering perturbations called \emph{mutations} over these extracted facts. %, illustrated Figure~\ref{fig:fever_construction}. 
Each claim was independently labeled using retrieved evidence.
Our reference corrections are the \emph{unmodified} facts extracted from Wikipedia.

The class balance and size of the dataset is reported in Table~\ref{tab:data:size}. 
The training and test splits are disjoint by entity. The additional hidden shared task test set was not used. 
%A portion of the claims in FEVER cannot be verified by evidence and are 
The claims labelled as \textsc{NotEnoughInfo}. 
%While these 
are used for training fact verification classifiers, but they will not be used for training the error correction systems in this paper as there is no labeled evidence to make corrections from. For completeness, we also release these unused \textsc{NotEnoughInfo} instances, as they have claims paired unmodified extracted facts (21934 training, 1870 development and 2037 test).

\begin{table}[h]
\centering
\footnotesize
\begin{tabular}{lccc}
\toprule
\multicolumn{1}{c}{\multirow{2}{*}{\textbf{Label}}} & \multicolumn{3}{c}{\textbf{Instance Count}}                                     \\ \cline{2-4} 
\multicolumn{1}{c}{}                                & \textbf{Train}            & \textbf{Validation}             & \textbf{Test}            \\ \midrule
Supports                                            & 37961                     & 1477                     & 1593                     \\
Refutes                                             & 20075                     & 2091                     & 2289                     \\ \midrule
Total Training                                         & \multicolumn{1}{c}{58036} & \multicolumn{1}{c}{3568} & \multicolumn{1}{c}{3891} \\
\bottomrule
\end{tabular}
\caption{Instance counts by class and dataset partitions}
\label{tab:data:size}
\end{table}

\section{Evaluation}
%Due to the difficulties in automated evaluation of generated language, 
While it's convenient to use an automatic metric during development, these metrics compute token overlap against a single reference sentence and cannot capture the nuances required to assess the veracity of the generated corrections against evidence. 
Thus, our primary evaluation will use human raters to label whether the model predictions meet the task requirements stated in Section~\ref{sec:task}. 

Human raters are asked three questions about system outputs to assess whether the corrections meet the requirements of intelligibility, supported by evidence, and error correction introduced in Section~\ref{sec:task}. 
For the first 2 requirements, the question has a binary answer. For the third requirement of error correction, the question has 3 answer choices: (1) the information content w.r.t.\ the evidence improved, (2) information unrelated to the claim was added (i.e. the claim was ignored), (3) no correction was needed (i.e.\ the claim was already supported by evidence).
%Annotators are asked three questions which had a binary answer for whether the correction is fluent and supported by evidence, and a three way answer for whether the error is corrected. 
%This additional neutral choice for the final question accounts for cases where an error is removed, but replaced with an alternative fact rather than a correction. 
The raters were shown each question in this sequence without knowledge of which system generated the correction. 
Negative answers to a question automatically assigned negative answers to subsequent ones (prescribing that an unintelligible sentence could not contain a fact supported by evidence or introduce a correction).
20\% of the tasks are assigned to two raters to measure inter-annotator agreement.
We used 4 expert participants from our lab (none of them co-authors of the paper) who were familiar with fact verification, but not with error correction. Responses were calibrated using a pilot study on the validation set.

For automated evaluation, we use SARI \citep{xu-etal-2016-optimizing} which is a metric used for sentence simplification. SARI considers ngrams retained from the source as well added or deleted ngrams through comparison against a reference sentence. We additionally report BLEU \cite{Papineni2002} and ROUGE \citep{Lin2004} to indicate precision and recall of the correction.
In Section~\ref{sec:results}, we report correlation of automated metrics against our manual evaluation.  %The results of this pilot study was used to select which automated metric (out of a choice of BLEU, ROUGE, SARI, BERTScore, NLI Label Flips) would be used for system development. 

% , these  We con sidered, bleu, rouge, sari popular in literature. While calculating all of them the main evaluation metric is SARI - it has nice properties that align with the requirement of the task. And, we will be verifying the choice of SARI with qualitative evaluation as. 

% Main results are human evaluation. SARI based evaluation- SARI was chosen using the development set using prototype experiments. Results in section 8 are reported on. 

\section{Implementation}
\label{sec:impl}

\paragraph{T5 Masker-Corrector}
We fine-tuned the T5-base pre-trained models released by HuggingFace \citep{wolf-etal-2020-transformers}.
The number of training epochs and learning rate was selected through optimizing the overall SARI score. The search space for learning rate was $\{10^{-5},5\cdot10^{-5},10^4,5\cdot10^{-4}\}$. We used $5\cdot 10^{-5}$ for all experiments. We found diminishing returns in SARI after 4 epochs and stopped training.

\paragraph{Fully Supervised Ceiling}
We use this model to estimate the ceiling performance of a factual error correction system (assuming a reasonable amount of training data is available) that other methods can be compared against. 
We fine-tune a T5-base model with supervision of the correction (see Section~\ref{sec:data}), using the same hyper-parameter choices as the T5 Masker-Corrector.

\paragraph{Automated Scoring} A single reference sentence from the FEVER dataset is used for automated scoring. We consider BLEU, ROUGE, and SARI. SARI considers the F1 of added tokens, F1 of kept tokens, precision of deletions, and the mean of these 3 scores (denoted \textit{final}). We use code made available by \citet{xu-etal-2016-optimizing}.

\paragraph{Evidence Retrieval} We use the Facebook implementation of DPR \citep{Karpukhin2020} without fine-tuning and constructed an index over the Wikipedia version released with FEVER \citep{Thorne2018a}, chunked into passages of 50 tokens. %To increase precision, the DPR results are filtered to keep those matching page-level predictions from GENRE \citep{DeCao2020AutoregressiveRetrieval}. 
For GENRE, the original authors' implementation was used.  We selected the top matching 2 passages. This resulted in the highest scores on the downstream corrections; SARI was lower when using $1$ or $3$ passages. 

\paragraph{Maskers} For the white-box masker, we use the implementation provided by \citet{Shah2019} applied to our dataset retaining original hyper-parameters trained on FEVER. For the black-box masker, we use the LIME implementation from \citep{Ribeiro2016} to probe a BERT classifier \citep{Devlin2019} fine-tuned on FEVER. For the LM and random baseline maskers, where the number of masks was tunable, we masked 50\% of the tokens, which was similar to the number of tokens masked by the black- and white-box maskers.

\paragraph{Language Model as Correctors?} We greedily decode masked tokens using a BERT-base-cased language model using the HuggingFace implementation \citep{wolf-etal-2020-transformers} without fine-tuning. 

\paragraph{Comparison to Previous Work} For comparison to previous work, we use the dual-encoder pointer network implementation from \cite{Shah2019}, retaining the original hyper-parameter choices.

% For maskers that queried a fact verification model, these models were trained using a variant of the FEVER dataset which sub-sampled \textsc{Supported} instances during training to ensure class balance and generated \textsc{NotEnoughInfo} instances by randomly sampling negative evidence for equal number of \textsc{Supported} and \textsc{Refuted} claims using evidence       retrieved from DPR.

%  against a BERT classifier \citep{Devlin2019} fine-tuned for 3 epochs, optimizing label accuracy. For the white-box masker, we use the ESIM model implementation released by \citet{Shah2019} with our updated dataset. To compare against the corrector released by \citet{Shah2019}, we also use their implementation with our data retaining the authors' hyper-parameter choices and training regimen.

\begin{table*}[t]
\centering
\begin{tabular}{@{}ccccccc@{}}
\toprule
 \multirow{2}{*}{\textbf{System}}  & \multirow{2}{*}{\textbf{Evidence}} & \multirow{2}{*}{\textbf{\begin{tabular}[c]{@{}c@{}}Training\\ Masks\end{tabular}}} & \multirow{2}{*}{\textbf{\begin{tabular}[c]{@{}c@{}}Test \\ Masks\end{tabular}}} & \multicolumn{3}{c}{\textbf{Aggregated Score (\%)}} \\ \cmidrule(l){5-7} 
 &  &  &  &  {\small\textbf{Intelligible}} & {\small\textbf{Supported}} & {\small\textbf{Corrected}} \\ \midrule
\multicolumn{1}{c}{T5 Fully Supervised} & Gold & - & -  & 98.9 & 88.9 & 68.9 \\
\multicolumn{1}{c}{T5 Fully Supervised} & Retrieved & - & -  & 97.7 & 64.7 & 48.9 \\ \midrule
T5 Masker + Corrector & Retrieved & Random & Heuristic  & 89.3 & \textbf{57.9} & \textbf{40.0} \\
T5 Masker + Corrector & Retrieved & Heuristic & Heuristic & 90.0 & 38.0 & 20.0 \\
T5 Masker + Corrector & Retrieved & Random & Black-box & \textbf{93.1} & 42.2 & 24.0 \\
T5 Masker + Corrector & Retrieved & Black-box & Black-box & 91.4 & 37.0 & 19.8 \\
T5 Masker + Corrector & Retrieved & White-box & White-box  & 90.6 & 41.7 & 23.9 \\
\midrule
BERT Language Model & -  &  - & Heuristic & 48.0 & 20.7 & 15.0 \\
BERT Language Model & -  &  - & Black-box & 30.1 & 4.9 & 3.4 \\
\citet{Shah2019} M+C & Gold & White-box & White-box  & 32.2 & 10.7 & 5.0 \\
% BERT LM & - & - & Black-box & 99.7 & 26.9 & 0.0 & 0.0 \\
%Masked LM & Greedy BERT LM decode & 100 & 32.9 & 7.4 &  4.0  \\ 
\bottomrule
\end{tabular}
\caption{Aggregated scores from human evaluation considering intelligibility, whether generated instances were supported by evidence and errors corrected.}
\label{tab:results:aggregate}
\end{table*}

\section{Results}
\label{sec:results}
\label{sec:results:human}

%Using retrieved evidence added noise, harming these scores as unrelated evidence was often retrieved. Using random masks when training the corrector resulted in a higher number of corrected claims than using a specific masker for training. For the DPR black-box model, this resulted in an improvement of the correction rate by 10\%. 
We first report results from a manual evaluation, assessing the requirements that corrections are intelligible, supported by evidence, and improve the factuality of the claim, as listed in Section~\ref{sec:task}. Our evaluation considers a sample of 200 instances per system. We report the results in Table~\ref{tab:results:aggregate}. For inter-annotator agreement control, 20\% of instances were annotated by two annotators: the Cohen's $\kappa$ scores for the 3 questions are 0.92 for intelligible, 0.92 for supported, and 0.86 for corrected. 
When using retrieved evidence, the white-box masker generated no masks for 41\% of instances. Without masked tokens, the T5 corrector copied the input claim to the output. This fits the assumption that, if the claim is already supported well by evidence, no correction is required. %change is required i. % all instances in the test set, reflected in the \emph{success rate} column. Where no masks were generated, the instances were not considered for scoring or manual evaluation.

The fully supervised models had the highest rate of satisfactory corrections that improved the factuality of the claim (requirement 3), indicating a performance ceiling for the distantly-supervised models. Incorporating retrieved evidence in these supervised models (rather than gold) reduced the number of corrections supported by evidence from 88.9\% to 64.7\% and the number of satisfactory corrections from 68.9\% to 48.9\% showing the challenges of incorporating (possibly noisy) retrieved evidence when generating the corrections. 

When using the masker and corrector distant supervision strategy, different maskers could be used to train the corrector to the masker used at test time. We observed that training the corrector with random masks yielded both a higher rate of satisfactory corrections and corrections supported by evidence when using either the black-box or heuristic masker at test time. We further evaluate other maskers with automated metrics in Section~\ref{sec:results-random}.

Using a  heuristic masker at test time, which removed tokens from the claim not present in the evidence, generated more claims meeting the supported and corrected requirements than masks generated by querying a fact verification model (both black-box and white-box). An analysis of the masker's influence on the corrections is provided in Section~\ref{sec:results-masker}. 
The two baseline systems, Dual Encoder M+C, based on \citet{Shah2019}, and a pre-trained BERT language model, %did not
generated corrections that were intelligible or supported by evidence at a lower rate than the aforementioned models, further discussed in Sections~\ref{sec:results-baseline} and~\ref{sec:results-lm}. 

We report the correlation between automated scoring metrics and our manual evaluation in Table~\ref{tab:correlation}. The \textsc{Keep} component of SARI, which measures the F1 of n-grams from the claim retained in the output, had the highest correlation with all three requirements. %, and very strong correlation with whether the claim was corrected. 
%and acts as useful diagnostic. 
Overly aggressive maskers which remove too much content from the claim can result in unintelligible outputs, or corrections unrelated to the claim. % This fits with the assumption that 
ROUGE2, which measures the recall of bigrams in the correction w.r.t.\ the reference, exhibited reasonable correlation to the manual evaluation against the supported and corrected requirements, however does not correlate as well with intelligibility. %as SARI \textsc{Keep}. 
The \textsc{Add} and \textsc{Delete} components of SARI provide further information but do not correlate as strongly with the human judgements. % as the other metrics we considered. 
% Both \textsc{Keep} and \textsc{Add} are precision-oriented
% %have a precision component 
% w.r.t\ the reference, which, as indicated by BLEU does not correlate well. %Limitations in , and 
Having only one reference correction reduces the utility of precision-oriented metrics, like BLEU, as valid corrections can %be made that 
differ from the reference.%  ever, with only one reference correction, their correlation with human judgements of whether the claim is corrected is lower -- indicating that 

%.  BLEU, which is precision based, did not correlate strongly with any of our requirements.  
\begin{table}[th!]
\footnotesize
\centering
\begin{tabular}{@{}cccc@{}}
\toprule
\multirow{2}{*}{\textbf{Metric}} & \multicolumn{3}{c}{\textbf{Correlation (Pearson r)}} \\ \cmidrule(l){2-4} 
 & \textbf{Intelligible} & \textbf{Supported} & \textbf{Corrected} \\ \midrule
SARI Keep & \textbf{.87} & \textbf{.95} & \textbf{.93} \\
SARI Final & .78 & .92 & .91\\
SARI Delete & .72 & .82 & .91 \\
SARI Add & .52 & .84 & .79 \\
ROUGE2 & .75 & .90 & .91 \\ %\textbf{.95} \\
ROUGE1 & .71 & .87 & .88 \\
BLEU2 & $-$.05 & .32 & .45 \\
BLEU1 & $-$.46 & $-$.10 & .05 \\
\bottomrule
\end{tabular}
\caption{Both SARI and ROUGE automated scoring metrics have high correlation to manual evaluation.}
%for the corrected claims. }
\label{tab:correlation}
\end{table}

% %In Ta, we report the correlation between the human evaluation and automated metrics. 
% Most automated evaluation metrics exhibited a high correlation against whether the correction is supported by evidence and correcting the error in the claim (\ref{tab:correlation}). However, BLEU did not correlate as strongly with either requirement.  

% XXX 
% To do : rewrite
% ROUGE and all SARI components correlated with whether the correction was supported by evidence and correcting any errors. However, SARI is the only scoring metric to consider the source claim for error correction. Correlation between these metrics and the fluency was lower in all cases.
% Even though the SARI Add scores were low for all systems (with the exception of full supervision), they still correlated well with the requirement 2 and 3 in the manual evaluation. In this manual evaluation, it was observed that claims can be corrected without the need to add new tokens. For example ``Donald Trump was president of France'' can be corrected by deleting ``of France'' without copying new tokens from the evidence. This is in contrast to the reference correction which may have used ``of the USA'' to undo a entity substitution introduced by the annotator when generating the original mutated claim.

\subsection{Choice of masker}
\label{sec:results-masker}
When training the corrector with the same masker that is used at test time, both the heuristic and black-box maskers yielded comparable scores under human evaluation. Inspection of SARI breakdown in Table~\ref{tab:results:masker} indicates that more tokens were kept when using the heuristic masker (Keep=.651) whereas the black box model was more aggressive in masking, resulting in less information from the claim being retained (Keep=.594). This correlated well with human judgements as more information retained gives a richer context for generating the correction and prevents erasure of claims already (partially) supported by the evidence.

Both the black-box (LIME) and white-box (the masker from \citet{Shah2019}) methods require querying a %fact
veracity classifier to generate the masks.
Using retrieved evidence for the veracity classifier, which was used to generate the masks in conjunction with these two methods, had a negative impact on most components of the SARI score. %when generating the masks had different effects for the black-box and white-box maskers.
For the black-box masker, using retrieved evidence reduced the number of masked tokens from an average of $4.7$ per claim to $3.9$. Whereas the number of masked tokens by the white-box masker remained unchanged at $4.7$ (approximately 50\% of number of tokens in the claim). 
Most notably, the white-box method of mask generation (row 4 in Table~\ref{tab:results:masker}) did not to generate masks for 41\% of instances when using retrieved evidence, whereas all instances had at least one mask when using gold evidence -- an artefact of the noise introduced by retrieval. %Despite this limitation,

\begin{table}[th!]
\centering
\footnotesize
\begin{tabular}{@{}cccccc@{}}
\toprule
\multirow{2}{*}{\textbf{Masker}} & \multicolumn{4}{c}{\textbf{SARI Score}} \\ \cmidrule(l){2-5} 
 &  \textbf{Keep} & \textbf{Delete} & \textbf{Add} & \textbf{Final} \\ \midrule
\begin{tabular}[c]{@{}l@{}}Black-box (Gold) \end{tabular} & .630 & \textbf{.582} & .088 & .433  \\
\begin{tabular}[c]{@{}l@{}}White-box (Gold) \end{tabular} &  \textbf{.652} & .559 & \textbf{.128} & \textbf{.447}\\
\begin{tabular}[c]{@{}l@{}}Black-box (IR)\end{tabular}&   .594 & .526 &.090 & .412 \\
\begin{tabular}[c]{@{}l@{}}White-box (IR)\end{tabular}&  .628 & .535 & .107 & .426 \\
\begin{tabular}[c]{@{}l@{}}Heuristic (IR)\end{tabular} &.651 & .574 & .041 & .422 \\
Masked LM & .538 & .509 & .062 & .370 \\
Random & .619 & .475 & .087 & .390 \\ \bottomrule
\end{tabular}
\caption{Extrinsic evaluation of maskers, varying the use of evidence when generating the masks, evaluated using the T5 Masker+Corrector model. }
\label{tab:results:masker}
\end{table}

\subsection{Corrector trained with random masks}
\label{sec:results-random}
Generating large quantities of masked training data through querying a model, such as with the black-box model explanation techniques, can be computationally expensive. In contrast, random masks can be generated without querying a model.
Using a corrector trained on random masks resulted in higher quality outputs at test time when paired the black-box and heuristic maskers. Training with random masks promotes good exploration of the task.
%and mitigates some effects of exposure bias.
In contrast, while the black-box and heuristic approaches worked well during testing, correctors \emph{trained} on these maskers generated worse outputs due to the limited exploration of the task space. Additionally, generating training data using the black- and white-box methods requires making predictions using the model's training data which may result in different outcomes to making predictions on unseen test data.
% This is because these maskers exploit model behavior and token overlap which do not explore enough of the task. 
% Random gives good exploration of the task space as. Heuristic is good to apply at test, but not training as it doesn't explore enough of the task.

% resulted in better judgements of human evaluation and is also reflected in automated results, which are reported in Table~\ref{tab:results:random}. There may be two reasons for this: the black- and white- box maskers exhibit exposure bias, as the masks generated on the training set overlap with the data the classifier is trained on, and random masking is not informed by the claim or evidence,
% resulting in a wider coverage of masking positions.% as it is not informed by the claim or evidence. 
% \begin{table}[h!]
% \begin{tabular}{@{}ccccc@{}}
% \toprule
% \multirow{2}{*}{\textbf{System}} & \multicolumn{4}{c}{\textbf{SARI Score (\%)}} \\
%  & \textbf{Add} & \textbf{Delete} & \textbf{Keep} & \textbf{Final} \\ \midrule
% NLI (LIME)+T5 &  &  &  &  \\
% LM (BERT)+T5 &  &  &  &  \\
% NLI (Attn)+T5 &  &  &  &  \\
% NLI (Attn)+Ptr &  &  &  &  \\ \bottomrule
% \end{tabular}
% \caption{Masker and corrector architectures. These systems su}
% \end{table}

\begin{table}[h!]
\centering
\footnotesize
\begin{tabular}{@{}cccccc@{}}
\toprule
\multirow{2}{*}{\textbf{Masker}} & \multicolumn{4}{c}{\textbf{SARI Score}} \\ \cmidrule(l){2-5} 
  & \textbf{Keep} & \textbf{Delete} & \textbf{Add} & \textbf{Final} \\ \midrule
\begin{tabular}[c]{@{}l@{}}Black-box (Gold) \end{tabular}  & .618 & .622 & .102 &.447  \\
\begin{tabular}[c]{@{}l@{}}White-box (Gold)\end{tabular}  & .640 & .570 & .114 & .441  \\
\begin{tabular}[c]{@{}l@{}}Black-box (IR)\end{tabular} & .611 & .543 & \textbf{.194} & .419 \\
\begin{tabular}[c]{@{}l@{}}White-box (IR)\end{tabular} & .618 & .590 & .144 & .452   \\
\begin{tabular}[c]{@{}l@{}}Heuristic (IR)\end{tabular}   & \textbf{.652} & \textbf{.627} & .155 & \textbf{.478}  \\
Masked LM & .561 & .529 & .078 & .389  \\
\bottomrule
\end{tabular}
\caption{Using random masks at training resulted in higher scores when testing with different maskers}
\label{tab:results:random}
\end{table}

\subsection{Comparison to previous work}
\label{sec:results-baseline}
% In Section~\ref{sec:related}, we identified that knowledge captured in language models can be used to make corrections in claims \citep{Petroni2019} and that pointer networks can be used to generate corrections using an independently encoded claim and evidence \citep{Shah2019}. For completeness, we include the outcomes of these systems in Table~\ref{tab:results:mlm} and Table~\ref{tab:results:corrector} respectively. 

Previous work uses a dual encoder pointer network \citep{Shah2019} to make corrections, reported in Table~\ref{tab:results:shah}. The corrector tended to copy portions of claim rather than correct it, resulting in a SARI \textsc{Keep} score of .452 which is lower than the T5 model using the same white-box masker (Table~\ref{tab:results:masker}). % This model used a limited vocabulary, 
Human evaluation considered these corrections mostly unintelligible, even when using gold evidence (Table~\ref{tab:results:aggregate}). %While the masks generated by this model had utility , 
This was especially the case for rarer entities. %Additionally, the corrector encoded the claim and evidence separately, limiting the output. 
Hyper-parameter tuning of the corrector's coverage ratio, as suggested by the authors, did not yield improvements. 
%We used the  using the system released by \citet{Shah2019} did not introduce many new, reflected in the low SARI Add score. However, the final SARI score is lower than the systems using the T5 corrector, as fewer tokens from the claim were kept, indicated by the lower SARI Keep score. While the protocol for white-box masking was useful when combined with the T5 transformer corrector, independently encoding the claim and evidence did not result in satisfactory corrections, especially when combining retrieved evidence in the corrector (Table~\ref{tab:results:corrector}, row 2).  

\begin{table}[h]
\centering
\footnotesize
\begin{tabular}{@{}cccccc@{}}
\toprule
\multirow{2}{*}{\textbf{System}} & \multicolumn{4}{c}{\textbf{SARI Score}} \\ \cmidrule(l){2-5} 
  & \textbf{Keep} & \textbf{Delete} & \textbf{Add} & \textbf{Final} \\ \midrule
Dual Enc Ptr (Gold) & .\textbf{452} & .\textbf{569} &  .\textbf{039} & .\textbf{353} \\
Dual Enc Ptr (IR) & .345 & .481 &  .017  & .281  \\ \bottomrule
\end{tabular}
\caption{Results using a dual encoder pointer network \citep{Shah2019} were low, despite the strong masker.}

\label{tab:results:shah}
\end{table}

\subsection{Language Models as Correctors?}
\label{sec:results-lm}
With the exception of the heuristic masker, using a pre-trained language model, without fine-tuning, to correct claims resulted in low SARI scores (Table~\ref{tab:results:mlm}). Without conditioning on the evidence, the correction is not related to the claim or supported by evidence to verify the claim, which is indicated by the low SARI Add scores which consider the precision of the added tokens. As these maskers deleted most tokens, retaining only stop-words, decoding most likely tokens without a prompt or context tokens resulted in unintelligible outputs. For the heuristic masker, more content words were retained yielding more intelligible outputs. However, these were not always supported by evidence, indicated in the human evaluation in Table~\ref{tab:results:aggregate}.

% This model has two limitations: firstly it does not condition the correction on the original claim or evidence, and secondly, multiple tokens are masked and the baseline model independently decodes the missing tokens sometimes resulting in ungrammatical outputs.
\begin{table}[t]
\footnotesize
\centering
\begin{tabular}{@{}cccccc@{}}
\toprule
\multirow{2}{*}{\textbf{Masker}} & \multicolumn{4}{c}{\textbf{SARI Score}} \\ \cmidrule(l){2-5} 
 &  \textbf{Keep} & \textbf{Delete} & \textbf{Add} & \textbf{Final} \\ \midrule
Masked LM & .360 & .472 & .019 & .289 \\
Heuristic (IR) & .\textbf{629} & .\textbf{651} & .\textbf{034} & .\textbf{438} \\
White-box (IR) & .232 & .446 & .005 & .228  \\ 
Black-box (IR) &  .364 & .003 & .001 & .122 \\
\bottomrule
\end{tabular}
\caption{Correcting claims using a language model does not condition the generation on evidence.}
\label{tab:results:mlm}
\end{table}

% \begin{table*}[t]
% \centering
% \begin{tabular}{@{}cccccc@{}}
% \toprule
% \multirow{2}{*}{\textbf{Masker}} & \multirow{2}{*}{\textbf{Corrector}} & \multirow{2}{*}{\textbf{\begin{tabular}[c]{@{}c@{}}Success \\ Rate (\%)\end{tabular}}} & \multicolumn{3}{c}{\textbf{Aggregated Score (\%)}} \\ \cmidrule(l){4-6} 
%  &  &  & \textbf{Fluency} & \textbf{Supported} & \textbf{Corrected} \\ \midrule
% \multicolumn{2}{c}{Fully supervised T5 (Gold)} & 100 & 95.1 & 91.8  & 88.5  \\
% \multicolumn{2}{c}{Fully supervised T5 (DPR)} &  100 & 100  & 87.3 & 84.1 \\
% White-box (DPR) & T5 white-box (DPR) & 73.9 & 98.7  & 22.4 & 17.1  \\
% Black-box (DPR) & T5 black-box (DPR) & 99.7 & 90.9& 39.4  & 22.7  \\
% Black-box (DPR) & T5 random (DPR) & 99.7 & 94.4 & 51.4 & 36.1  \\
% White-box (Gold) & Dual Enc Pointer (Gold) & 100 &  67.7 & 16.1 & 0.8  \\
% Masked LM & BERT MLM & 100 & 32.2 & 6.8 & 3.4  \\ \bottomrule
% \end{tabular}
% \caption{Aggregated scores from human evaluation from a sample of 500 instances}
% \end{table*}

% The results, listed in Table~\ref{tab:results:mutation}, indicate that adding the mutation type information generated corrections with a moderately higher SARI score with increases to the Keep and Add components. While some mutation types provide information about which entities to substitute (for example, the generalization of pianist could be a musician), there is still ambiguity about which entities may have been substituted by the annotators when generating the mutations for extracted facts.

% JT Re-read all from here again
\section{Qualitative Error Analysis}
In this section we discuss the following issues which were present in all master-corrector systems: 

\paragraph{Over-erasure} In some instances, the masker removed most or all of the non-stopword tokens from the claim. This resulted in the original meaning of the claim being erased. Without this information the corrector could not reconstruct the claim, resulting in corrections that were unrelated to the input claim. 
%n't faithful to the input. 
This issue was most prevalent for the black-box masker, where 15\% of instances had more than 5 consecutive tokens masked and 32\% of instances had 4 consecutive tokens masked. In contrast, the heuristic masker, which identifies the tokens not present in the 
%performed a diff between the claim and
retrieved evidence
had 5 consecutive tokens masked for 3\% of instances and 4 consecutive tokens masked for 9\% of instances. 
While, in some cases, appropriate corrections could be made despite the aggressive masking (e.g.\ the claim ``Exit the King is by man[sic].''  was fully masked, but corrected to include the author's name), others were re-written focusing on a different fact, e.g.\ a claim about the length of reign of Maria Theresa was rewritten to be about her date of birth. % because all the tokens were masked out and the evidence contained both facts.

%\paragraph{Recovering from incorrect masks}
\paragraph{Incorrect masking} When the erroneous tokens in a claim were not masked, the corrector would generate outputs not supported by evidence. For example the following claim, which has an incorrect year, was masked but retaining the error: ``Ghost, the film was released in 1994''
as ``[MASK] , [MASK] [MASK] [MASK] [MASK] [MASK] in 1994''. Even with suitable retrieved evidence, indicating the release year is 1990, no appropriate correction could be made. % was made: ``Ghost, by the way, was released in 1994''.

\paragraph{Inadequate evidence retrieval} %Evidence is required when generating both the masks and corrections. Retrieving evidence that neither supported or refuted claims often caused the model to generate spurious masks as it was not possible for the parts needed correction to be identified. %The black-box masker was more sensitive to unrelated evidence and generated 5 or more consecutive masks for 15\% of instances whereas  generated  This was evident in the black-box masker which which generated 580 
%and resulted in spurious corrections. Inspection of instances with 5 or more consecutive masks indicates that the black-box model, where LIME explanations were used 
Where the evidence retrieved was related, but not specifically supporting or refuting the claim, the generated corrections were vague: the claim ``Poldark aired on HBO'' was corrected to ``Poldark premiered on TV'' as the evidence lacked the name of the correct TV station. Similarly, where incorrect masks were made, additional retrieval retrieval may be required to prevent the corrector from hallucinating information to cover the knowledge missing from the evidence. For example, the name of the TV show was masked in the claim ``Two and a half men starred Jamie Fox[sic]'', but as no mention of Jamie Fox was present in the evidence, the model hallucinated a different TV show name. % tokens were unrelated to   When retrieved evidence was about the  couldn't correct the error 
% Evidence didn't state that show was broadcast on HBO

\section{Conclusions and Future Work}
Going beyond simply identifying errors, factual error correction presents a number of %interesting research 
challenges for information retrieval, fact verification and abstractive summarization communities alike. In this paper, we demonstrated that the task can be performed with %both full supervision and 
distant supervision in the form of claims labeled by evidence supporting or refuting them.
%with reasonable success using existing technologies. 
However, there are a number of outstanding challenges that must be addressed.
%, including how evidence is incorporated into the correction and how the generated corrections are scored. 
The data we used from the FEVER task was re-purposed to evaluate whether systems can undo mutations introduced by human annotators and may not be representative of the range of factual errors that would be present in real-world documents. %Furthermore, the use of the extracted facts as references for scoring assesses whether systems can undo mutations introduced by the annotators rather than correct errors, warranting further manual evaluation.
While some automated metrics correlated well with human judgements,
%(with the exception of fluency),
future work should consider how automated scoring can be better used to discriminate the adequacy %and utility
of the generated corrections going beyond similarity to the reference sentence. 
From a modelling perspective, the masks strongly influenced the corrector and further work is required to generate masks that result in better corrections. We observed where masks mismatched the evidence, the correction was vague, hallucinated or did not correct the factual errors in the claim. This could be addressed through joint training of both components to enable them to avoid error propagation from masking to correction.

\section*{Acknowledgements}
The authors wish to thank: {Tal~Schuster} for his helpful comments and feedback; {Nicola~De~Cao} for providing the GENRE predictions for FEVER; {Amrith~Krishna}, {Guy~Aglionby}, {Rami~Aly} and {Zhijiang Guo} for manual evaluation of the model predictions. This research was supported by donation of compute resources from Google Cloud.
 James Thorne is supported by an Amazon Alexa Graduate Research Fellowship. Andreas Vlachos is supported by the ERC grant AVeriTeC (GA 865958).
 
%This is allowed on page 9
\section*{Broader Impact Statement}
Our experiments were performed on publicly available data about common facts from Wikipedia. These data are released under a creative-commons license.  The expert raters from our lab who manually reviewed the generated instances were volunteers and were compensated through quid-pro-quo help on their own projects.

The intended use of this project is to help explain reasoning using evidence, going beyond single-label classification. This adds an additional safeguard, making the decision process more transparent as poor predictions by our model expose limitations that would be hidden by classification. Our data is synthetic in nature and is biased towards synthetic facts from popular entities. 
Application to political or scientific domains would require additional work. Misinformation about populations that are under-represented in our data may not be accurately identified or corrected without further mitigation. One positive finding in our paper was that some of biases perpetuated in the hallucinations of language models were mitigated when conditioning the generation on retrieved evidence. 

Model fine-tuning took approximately 2 hours per experiment on a single P100 GPU. Generating LIME explanations of the training dataset took approximately one day -- motivating our experiments that used models trained on random or heuristic maskers which required fewer resources by several orders of magnitude.

\bibliographystyle{acl_natbib}
\bibliography{fixed_Refs}

\begin{thebibliography}{37}
\expandafter\ifx\csname natexlab\endcsname\relax\def\natexlab#1{#1}\fi

\bibitem[{Atanasova et~al.(2020)Atanasova, Simonsen, Lioma, and
  Augenstein}]{Atanasova2020GeneratingExplanations}
Pepa Atanasova, Jakob~Grue Simonsen, Christina Lioma, and Isabelle Augenstein.
  2020.
\newblock \href {https://doi.org/10.18653/v1/2020.acl-main.656} {Generating
  fact checking explanations}.
\newblock In \emph{Proceedings of the 58th Annual Meeting of the Association
  for Computational Linguistics}, pages 7352--7364. Association for
  Computational Linguistics.

\bibitem[{Augenstein et~al.(2019)Augenstein, Lioma, Wang, Chaves~Lima, Hansen,
  Hansen, and Simonsen}]{Augenstein2019}
Isabelle Augenstein, Christina Lioma, Dongsheng Wang, Lucas Chaves~Lima, Casper
  Hansen, Christian Hansen, and Jakob~Grue Simonsen. 2019.
\newblock \href {https://doi.org/10.18653/v1/D19-1475} {{M}ulti{FC}: A
  real-world multi-domain dataset for evidence-based fact checking of claims}.
\newblock In \emph{Proceedings of the 2019 Conference on Empirical Methods in
  Natural Language Processing and the 9th International Joint Conference on
  Natural Language Processing (EMNLP-IJCNLP)}, pages 4685--4697, Hong Kong,
  China. Association for Computational Linguistics.

\bibitem[{Cao et~al.(2020)Cao, Dong, Wu, and Cheung}]{Cao2020FactualModels}
Meng Cao, Yue Dong, Jiapeng Wu, and Jackie Chi~Kit Cheung. 2020.
\newblock \href {https://doi.org/10.18653/v1/2020.emnlp-main.506} {{Factual
  Error Correction for Abstractive Summarization Models}}.
\newblock In \emph{Empirical Methods in Natural Language Processing}, pages
  6251--6258.

\bibitem[{Cao et~al.(2021)Cao, Izacard, Riedel, and
  Petroni}]{DeCao2020AutoregressiveRetrieval}
Nicola~De Cao, Gautier Izacard, Sebastian Riedel, and Fabio Petroni. 2021.
\newblock \href {https://openreview.net/forum?id=5k8F6UU39V} {Autoregressive
  entity retrieval}.
\newblock In \emph{International Conference on Learning Representations}.

\bibitem[{Chen et~al.(2017)Chen, Zhu, Ling, Wei, Jiang, and Inkpen}]{Chen2016}
Qian Chen, Xiaodan Zhu, Zhen-Hua Ling, Si~Wei, Hui Jiang, and Diana Inkpen.
  2017.
\newblock \href {https://doi.org/10.18653/v1/P17-1152} {Enhanced {LSTM} for
  natural language inference}.
\newblock In \emph{Proceedings of the 55th Annual Meeting of the Association
  for Computational Linguistics (Volume 1: Long Papers)}, pages 1657--1668,
  Vancouver, Canada. Association for Computational Linguistics.

\bibitem[{Cohen et~al.(2011)Cohen, Li, Yang, and Yu}]{Cohen2011}
Sarah Cohen, Chengkai Li, Jun Yang, and Cong Yu. 2011.
\newblock \href
  {http://static.googleusercontent.com/external_content/untrusted_dlcp/research.google.com/en//pubs/archive/37381.pdf}
  {{Computational Journalism: a call to arms to database researchers}}.
\newblock \emph{Proceedings of the 5th Biennial Conference on Innovative Data
  Systems Research (CIDR 2011) Asilomar, California, USA.}, (January):148--151.

\bibitem[{Devlin et~al.(2019)Devlin, Chang, Lee, and Toutanova}]{Devlin2019}
Jacob Devlin, Ming-Wei Chang, Kenton Lee, and Kristina Toutanova. 2019.
\newblock \href {https://doi.org/10.18653/v1/N19-1423} {{BERT}: Pre-training of
  deep bidirectional transformers for language understanding}.
\newblock In \emph{Proceedings of the 2019 Conference of the North {A}merican
  Chapter of the Association for Computational Linguistics: Human Language
  Technologies, Volume 1 (Long and Short Papers)}, pages 4171--4186,
  Minneapolis, Minnesota. Association for Computational Linguistics.

\bibitem[{Guu et~al.(2020)Guu, Lee, Tung, Pasupat, and Chang}]{Guu2020}
Kelvin Guu, Kenton Lee, Zora Tung, Panupong Pasupat, and Ming-wei Chang. 2020.
\newblock \href {https://arxiv.org/abs/2002.08909} {{REALM: Retrieval-Augmented
  Language Model Pre-Training}}.

\bibitem[{Han et~al.(2010)Han, Tetreault, Lee, and Ha}]{han-etal-2010-using}
Na-Rae Han, Joel Tetreault, Soo-Hwa Lee, and Jin-Young Ha. 2010.
\newblock \href
  {http://www.lrec-conf.org/proceedings/lrec2010/pdf/821_Paper.pdf} {Using an
  error-annotated learner corpus to develop an {ESL}/{EFL} error correction
  system}.
\newblock In \emph{Proceedings of the Seventh International Conference on
  Language Resources and Evaluation ({LREC}'10)}, Valletta, Malta. European
  Language Resources Association (ELRA).

\bibitem[{Hochreiter and Schmidhuber(1997)}]{Hochreiter1997}
Sepp Hochreiter and Jurgen Schmidhuber. 1997.
\newblock \href {https://doi.org/10.1162/neco.1997.9.8.1735} {{Long Short-Term
  Memory}}.
\newblock \emph{Neural Computation}, 9(8):1735--1780.

\bibitem[{Holtzman et~al.(2020)Holtzman, Buys, Du, Forbes, and
  Choi}]{Holtzman2019TheDegeneration}
Ari Holtzman, Jan Buys, Li~Du, Maxwell Forbes, and Yejin Choi. 2020.
\newblock \href {https://openreview.net/forum?id=rygGQyrFvH} {The curious case
  of neural text degeneration}.
\newblock In \emph{International Conference on Learning Representations}.

\bibitem[{Karadzhov et~al.(2017)Karadzhov, Nakov, M{\`a}rquez,
  Barr{\'o}n-Cede{\~n}o, and Koychev}]{Karadzhov2017FullySources}
Georgi Karadzhov, Preslav Nakov, Llu{\'\i}s M{\`a}rquez, Alberto
  Barr{\'o}n-Cede{\~n}o, and Ivan Koychev. 2017.
\newblock \href {https://doi.org/10.26615/978-954-452-049-6_046} {Fully
  automated fact checking using external sources}.
\newblock In \emph{Proceedings of the International Conference Recent Advances
  in Natural Language Processing, {RANLP} 2017}, pages 344--353. INCOMA Ltd.

\bibitem[{Karpukhin et~al.(2020)Karpukhin, O{\u{g}}uz, Min, Lewis, Wu, Edunov,
  Chen, and Yih}]{Karpukhin2020}
Vladimir Karpukhin, Barlas O{\u{g}}uz, Sewon Min, Patrick Lewis, Ledell Wu,
  Sergey Edunov, Danqi Chen, and Wen-tau Yih. 2020.
\newblock \href {http://arxiv.org/abs/2004.04906} {{Dense Passage Retrieval for
  Open-Domain Question Answering}}.

\bibitem[{Knight and Chander(1994)}]{Knight1994AutomatedDocuments}
Kevin Knight and Ishwar Chander. 1994.
\newblock \href {https://www.aaai.org/Papers/AAAI/1994/AAAI94-119.pdf}
  {{Automated postediting of documents}}.
\newblock In \emph{Proceedings of the National Conference on Artificial
  Intelligence}, volume~1, pages 779--784.

\bibitem[{Kotonya and Toni(2020)}]{Kotonya2020}
Neema Kotonya and Francesca Toni. 2020.
\newblock \href {http://arxiv.org/abs/2010.09926} {{Explainable Automated
  Fact-Checking for Public Health Claims}}.
\newblock In \emph{The 2020 Conference on Empirical Methods in Natural Language
  Processing}.

\bibitem[{Lee et~al.(2020)Lee, Li, Wang, Yih, Ma, and
  Khabsa}]{Lee2020LanguageCheckers}
Nayeon Lee, Belinda Li, Sinong Wang, Wen-tau Yih, Hao Ma, and Madian Khabsa.
  2020.
\newblock \href {https://doi.org/10.18653/v1/2020.fever-1.5} {Language models
  as fact checkers?}
\newblock In \emph{Proceedings of the Third Workshop on Fact Extraction and
  VERification (FEVER)}, volume~2, pages 36--41. Association for Computational
  Linguistics.

\bibitem[{Lewis et~al.(2020)Lewis, Perez, Piktus, Petroni, Karpukhin, Goyal,
  K{\"{u}}ttler, Lewis, Yih, Rockt{\"{a}}schel, Riedel, and Kiela}]{Lewis2020}
Patrick Lewis, Ethan Perez, Aleksandara Piktus, Fabio Petroni, Vladimir
  Karpukhin, Naman Goyal, Heinrich K{\"{u}}ttler, Mike Lewis, Wen-tau Yih, Tim
  Rockt{\"{a}}schel, Sebastian Riedel, and Douwe Kiela. 2020.
\newblock \href {http://arxiv.org/abs/2005.11401} {{Retrieval-Augmented
  Generation for Knowledge-Intensive NLP Tasks}}.

\bibitem[{Lin(2004)}]{Lin2004}
Chin-Yew Lin. 2004.
\newblock \href {https://doi.org/10.1253/jcj.34.1213} {{ROUGE}: A package for
  automatic evaluation of summaries}.
\newblock In \emph{Text Summarization Branches Out}, pages 74--81, Barcelona,
  Spain. Association for Computational Linguistics.

\bibitem[{Ng et~al.(2014)Ng, Wu, Briscoe, Hadiwinoto, Susanto, and
  Bryant}]{Ng2014}
Hwee~Tou Ng, Siew~Mei Wu, Ted Briscoe, Christian Hadiwinoto, Raymond~Hendy
  Susanto, and Christopher Bryant. 2014.
\newblock \href {https://doi.org/10.3115/v1/W14-1701} {The {C}o{NLL}-2014
  shared task on grammatical error correction}.
\newblock In \emph{Proceedings of the Eighteenth Conference on Computational
  Natural Language Learning: Shared Task}, July, pages 1--14. Association for
  Computational Linguistics.

\bibitem[{Nie et~al.(2019)Nie, Yao, Wang, Pan, and Lin}]{nie-etal-2019-simple}
Feng Nie, Jin-Ge Yao, Jinpeng Wang, Rong Pan, and Chin-Yew Lin. 2019.
\newblock \href {https://doi.org/10.18653/v1/P19-1256} {A simple recipe towards
  reducing hallucination in neural surface realisation}.
\newblock In \emph{Proceedings of the 57th Annual Meeting of the Association
  for Computational Linguistics}, pages 2673--2679, Florence, Italy.
  Association for Computational Linguistics.

\bibitem[{Papineni et~al.(2002)Papineni, Roukos, Ward, and Zhu}]{Papineni2002}
Kishore Papineni, Salim Roukos, Todd Ward, and Wei-Jing Zhu. 2002.
\newblock \href {https://doi.org/10.3115/1073083.1073135} {{B}leu: a method for
  automatic evaluation of machine translation}.
\newblock In \emph{Proceedings of the 40th Annual Meeting of the Association
  for Computational Linguistics}, July, pages 311--318. Association for
  Computational Linguistics.

\bibitem[{Petroni et~al.(2020)Petroni, Piktus, Fan, Lewis, Yazdani, De~Cao,
  Thorne, Jernite, Plachouras, Rockt{\"{a}}schel, and
  Riedel}]{Petroni2020KILT:Tasks}
Fabio Petroni, Aleksandra Piktus, Angela Fan, Patrick Lewis, Majid Yazdani,
  Nicola De~Cao, James Thorne, Yacine Jernite, Vassilis Plachouras, Tim
  Rockt{\"{a}}schel, and Sebastian Riedel. 2020.
\newblock \href {http://arxiv.org/abs/2009.02252} {{KILT: a Benchmark for
  Knowledge Intensive Language Tasks}}.

\bibitem[{Petroni et~al.(2019)Petroni, Rockt{\"a}schel, Riedel, Lewis, Bakhtin,
  Wu, and Miller}]{Petroni2019}
Fabio Petroni, Tim Rockt{\"a}schel, Sebastian Riedel, Patrick Lewis, Anton
  Bakhtin, Yuxiang Wu, and Alexander Miller. 2019.
\newblock \href {https://doi.org/10.18653/v1/D19-1250} {Language models as
  knowledge bases?}
\newblock In \emph{Proceedings of the 2019 Conference on Empirical Methods in
  Natural Language Processing and the 9th International Joint Conference on
  Natural Language Processing (EMNLP-IJCNLP)}, pages 2463--2473, Hong Kong,
  China. Association for Computational Linguistics.

\bibitem[{Raffel et~al.(2020)Raffel, Shazeer, Roberts, Lee, Narang, Matena,
  Zhou, Li, and Liu}]{Raffel2020}
Colin Raffel, Noam Shazeer, Adam Roberts, Katherine Lee, Sharan Narang, Michael
  Matena, Yanqi Zhou, Wei Li, and Peter~J. Liu. 2020.
\newblock \href {http://arxiv.org/abs/1910.10683} {{Exploring the Limits of
  Transfer Learning with a Unified Text-to-Text Transformer}}.
\newblock \emph{Journal of Machine Learning Research}, 21:1--67.

\bibitem[{Ribeiro et~al.(2016)Ribeiro, Singh, and Guestrin}]{Ribeiro2016}
Marco Ribeiro, Sameer Singh, and Carlos Guestrin. 2016.
\newblock \href {https://doi.org/10.18653/v1/N16-3020} {{``}why should {I}
  trust you?{''}: Explaining the predictions of any classifier}.
\newblock In \emph{Proceedings of the 2016 Conference of the North {A}merican
  Chapter of the Association for Computational Linguistics: Demonstrations},
  volume~39, pages 97--101. Association for Computational Linguistics.

\bibitem[{Rohrbach et~al.(2018)Rohrbach, Hendricks, Burns, Darrell, and
  Saenko}]{rohrbach-etal-2018-object}
Anna Rohrbach, Lisa~Anne Hendricks, Kaylee Burns, Trevor Darrell, and Kate
  Saenko. 2018.
\newblock \href {https://doi.org/10.18653/v1/D18-1437} {Object hallucination in
  image captioning}.
\newblock In \emph{Proceedings of the 2018 Conference on Empirical Methods in
  Natural Language Processing}, pages 4035--4045, Brussels, Belgium.
  Association for Computational Linguistics.

\bibitem[{See et~al.(2017)See, Liu, and Manning}]{See2017GetNetworks}
Abigail See, Peter~J. Liu, and Christopher~D. Manning. 2017.
\newblock \href {https://doi.org/10.18653/v1/P17-1099} {Get to the point:
  Summarization with pointer-generator networks}.
\newblock In \emph{Proceedings of the 55th Annual Meeting of the Association
  for Computational Linguistics (Volume 1: Long Papers)}, volume~1, pages
  1073--1083. Association for Computational Linguistics.

\bibitem[{Shah et~al.(2020)Shah, Schuster, and Barzilay}]{Shah2019}
Darsh~J Shah, Tal Schuster, and Regina Barzilay. 2020.
\newblock \href {http://arxiv.org/abs/1909.13838} {{Automatic Fact-guided
  Sentence Modification}}.
\newblock In \emph{Proceedings of the AAAI Conference on Artificial
  Intelligence}.

\bibitem[{Stammbach and Ash(2020)}]{Stammbach2020}
Dominik Stammbach and Elliott Ash. 2020.
\newblock {e-FEVER: Explanations and Summaries for Automated Fact Checking}.
\newblock In \emph{Proceedings of the 2020 Truth and Trust Online Conference
  (TTO 2020)}, page~32. Hacks Hackers.

\bibitem[{Taylor(1953)}]{Taylor1953ClozeReadability}
Wilson~L. Taylor. 1953.
\newblock \href {https://doi.org/10.1177/107769905303000401} {{“Cloze
  Procedure”: A New Tool for Measuring Readability}}.
\newblock \emph{Journalism Quarterly}, 30(4):415--433.

\bibitem[{Thorne et~al.(2018)Thorne, Vlachos, Christodoulopoulos, and
  Mittal}]{Thorne2018a}
James Thorne, Andreas Vlachos, Christos Christodoulopoulos, and Arpit Mittal.
  2018.
\newblock \href {https://doi.org/10.18653/v1/N18-1074} {{FEVER}: a large-scale
  dataset for fact extraction and {VER}ification}.
\newblock In \emph{Proceedings of the 2018 Conference of the North {A}merican
  Chapter of the Association for Computational Linguistics: Human Language
  Technologies, Volume 1 (Long Papers)}, pages 809--819, New Orleans,
  Louisiana. Association for Computational Linguistics.

\bibitem[{Vaswani et~al.(2017)Vaswani, Shazeer, Parmar, Uszkoreit, Jones,
  Gomez, Kaiser, and Polosukhin}]{Vaswani2017}
Ashish Vaswani, Noam Shazeer, Niki Parmar, Jakob Uszkoreit, Lilon Jones, Aidan
  Gomez, Łukasz Kaiser, and Illia Polosukhin. 2017.
\newblock \href {https://doi.org/10.1017/S0140525X16001837} {{Attention is all
  you need}}.
\newblock In \emph{31st Conference on Neural Information Processing Systems
  (NIPS 2017)}, Long Beach, CA, USA.

\bibitem[{Wadden et~al.(2020)Wadden, Lin, Lo, Wang, van Zuylen, Cohan, and
  Hajishirzi}]{Wadden2020FactClaims}
David Wadden, Shanchuan Lin, Kyle Lo, Lucy~Lu Wang, Madeleine van Zuylen, Arman
  Cohan, and Hannaneh Hajishirzi. 2020.
\newblock \href {http://arxiv.org/abs/2004.14974} {{Fact or Fiction: Verifying
  Scientific Claims}}.

\bibitem[{Wang(2017)}]{Wang2017a}
William~Yang Wang. 2017.
\newblock \href {https://doi.org/10.18653/v1/P17-2067} {{``}liar, liar pants on
  fire{''}: A new benchmark dataset for fake news detection}.
\newblock In \emph{Proceedings of the 55th Annual Meeting of the Association
  for Computational Linguistics (Volume 2: Short Papers)}, pages 422--426.
  Association for Computational Linguistics.

\bibitem[{Wolf et~al.(2020)Wolf, Debut, Sanh, Chaumond, Delangue, Moi, Cistac,
  Rault, Louf, Funtowicz, Davison, Shleifer, von Platen, Ma, Jernite, Plu, Xu,
  Le~Scao, Gugger, Drame, Lhoest, and Rush}]{wolf-etal-2020-transformers}
Thomas Wolf, Lysandre Debut, Victor Sanh, Julien Chaumond, Clement Delangue,
  Anthony Moi, Pierric Cistac, Tim Rault, Remi Louf, Morgan Funtowicz, Joe
  Davison, Sam Shleifer, Patrick von Platen, Clara Ma, Yacine Jernite, Julien
  Plu, Canwen Xu, Teven Le~Scao, Sylvain Gugger, Mariama Drame, Quentin Lhoest,
  and Alexander Rush. 2020.
\newblock \href {https://doi.org/10.18653/v1/2020.emnlp-demos.6}
  {{Transformers: State-of-the-Art Natural Language Processing}}.
\newblock In \emph{Proceedings of the 2020 Conference on Empirical Methods in
  Natural Language Processing: System Demonstrations}, pages 38--45, Online.
  Association for Computational Linguistics.

\bibitem[{Xu et~al.(2016)Xu, Napoles, Pavlick, Chen, and
  Callison-Burch}]{xu-etal-2016-optimizing}
Wei Xu, Courtney Napoles, Ellie Pavlick, Quanze Chen, and Chris Callison-Burch.
  2016.
\newblock \href {https://doi.org/10.1162/tacl_a_00107} {Optimizing statistical
  machine translation for text simplification}.
\newblock \emph{Transactions of the Association for Computational Linguistics},
  4:401--415.

\bibitem[{Zhou et~al.(2020)Zhou, Gu, Diab, Guzman, Zettlemoyer, and
  Ghazvininejad}]{Zhou2020DetectingGeneration}
Chunting Zhou, Jiatao Gu, Mona Diab, Paco Guzman, Luke Zettlemoyer, and Marjan
  Ghazvininejad. 2020.
\newblock \href {http://arxiv.org/abs/2011.02593} {{Detecting Hallucinated
  Content in Conditional Neural Sequence Generation}}.
\newblock pages 1--21.

\end{thebibliography}

\end{document}